\documentclass[letterpaper, 10 pt, conference]{ieeeconf}  
\usepackage[noadjust]{cite}
\usepackage[hyphens]{url}
\usepackage{booktabs}
\usepackage{multirow}
\usepackage{xcolor}
\usepackage{graphicx}
\usepackage{amssymb}
\usepackage{adjustbox}
\usepackage{subcaption}
\usepackage{comment}
\usepackage{siunitx}
\usepackage [english]{babel}
\usepackage [autostyle, english = american]{csquotes}
\MakeOuterQuote{"}

\IEEEoverridecommandlockouts                              

\title{\LARGE \bf
Surgical fine-tuning for Grape Bunch Segmentation \\ under Visual Domain Shifts
}

\author{Agnese Chiatti$^{1}$, Riccardo Bertoglio$^{1}$, Nico Catalano$^{1}$, Matteo Gatti$^{2}$, and Matteo Matteucci$^{1}$
\thanks{$^{1}$ Department of Electronics, Information and Bioengineering (DEIB), Politecnico di Milano, Milan, Italy {\tt\small \{name.surname\}@polimi.it}}%
\thanks{$^{2}$ Department of Sustainable Crop Production, Università Cattolica del Sacro Cuore, Piacenza, Italy {\tt\small matteo.gatti@unicatt.it}}%
}

\usepackage{tikz}
\usepackage{textcomp}
\usepackage{hyperref}
\usepackage{lipsum}

\newcommand\copyrighttext{%
  \footnotesize \textcopyright 2023 IEEE. 
  Personal use of this material is permitted. Permission from IEEE must be obtained for all other uses, in any current or future media, including reprinting/republishing this material for advertising or promotional purposes, creating new collective works, for resale or redistribution to servers or lists, or reuse of any copyrighted component of this work in other works.}
\newcommand\copyrightnotice{%
\begin{tikzpicture}[remember picture,overlay]
\node[anchor=south,yshift=10pt] at (current page.south) {\fbox{\parbox{\dimexpr\textwidth-\fboxsep-\fboxrule\relax}{\copyrighttext}}};
\end{tikzpicture}%
}

\begin{document}

\maketitle
\copyrightnotice
\thispagestyle{empty}
\pagestyle{empty}

\begin{abstract}
Mobile robots will play a crucial role in the transition towards sustainable agriculture. To autonomously and effectively monitor the state of plants, robots ought to be equipped with visual perception capabilities that are robust to the rapid changes that characterise agricultural settings. In this paper, we focus on the challenging task of segmenting grape bunches from images collected by mobile robots in vineyards. In this context, we present the first study that applies surgical fine-tuning to instance segmentation tasks. We show how selectively tuning only specific model layers can support the adaptation of pre-trained Deep Learning models to newly-collected grape images that introduce visual domain shifts, while also substantially reducing the number of tuned parameters.
\end{abstract}

\section{INTRODUCTION and BACKGROUND}

The climate change crisis has highlighted the importance of increasing the sustainability of food production, as prescribed in the European Commission's "Farm to Fork" strategy\footnote{\url{https://food.ec.europa.eu/horizontal-topics/farm-fork-strategy_en}}. In this regard, digital technologies are playing a crucial role in reducing the amount of water and chemicals used in agriculture \cite{bertoglio2021digital}. One of the key applications of digital technologies is the deployment of mobile robots, which can perform a range of tasks such as plant spraying \cite{li2022design}, weeding \cite{bertoglio2023comparative}, and harvesting \cite{kootstra2021selective}. To carry out these tasks effectively, robots need the ability to autonomously monitor plant traits and status, a task also known as \textit{plant phenotyping}. For example, in vineyards, a robot must be capable of detecting plant organs for posing the appropriate cuts during winter pruning operations \cite{guadagna2023using}. They also ought to accurately identify the presence of grape bunches, their level of ripeness, and promptly detect the emergence of any diseases that may compromise the fruit quality.

Robot's perception systems deployed in agricultural settings face particular challenges due to the significant weather and seasonal variations that characterise these environments. Thus, ensuring the effective reuse of visual patterns and features learned under specific environmental conditions (e.g., in terms of weather, lighting, and plant diversity) becomes crucial. This requirement stems from the need to guarantee accurate plant monitoring, even when the underlying conditions change. For instance, viewpoint changes caused by different sensor positions and occlusions caused by leaves are prominent factors that can hinder the accurate monitoring of fruit \cite{zaenker2021combining,tang2023optimization}.

The widespread application of Deep Learning (DL) methods has considerably accelerated the progress in various visual perception tasks, including plant phenotyping \cite{saleem2021automation}. However, supervised DL methods typically require abundant training data and are susceptible to changes in the data distribution. Moreover, training all model parameters on new data is a costly process in terms of computational power and memory footprint, especially when working on edge devices and mobile platforms. To address these issues, one possible approach is to pre-train the model on a large-scale source domain and fine-tune the parameters on a few examples from the target domain. The aim of fine-tuning is to adapt the model to the target domain while retaining the information learned during pre-training, particularly in cases where the source and target distributions significantly overlap despite the shift. This process is commonly known as transfer learning. A traditional transfer learning practice known as linear probing involves fine-tuning only the last few layers of a Deep Neural Network (DNN) while reusing features from earlier layers. This approach was based on initial evidence suggesting that representations in earlier layers may be more transferable to new data and tasks than the specialised features learned in higher layers \cite{yosinski2014transferable}.

Recent research \cite{lee2023surgical,evci2022head2toe} has explored effective alternatives to this consolidated fine-tuning practice. Indeed, Lee et al. \cite{lee2023surgical} discovered that selectively tuning only the earlier, intermediate, or last layers of a DNN can counteract different types of distribution shifts and often even outperform cases where all model parameters are tuned. They have named this approach \textit{surgical fine-tuning} (SFT). Their study concerned transfer learning across different image classification benchmarks, such as CIFAR and ImageNet. However, the authors' conclusions have yet to be validated on image segmentation tasks and data gathered in real-world application scenarios, e.g., from mobile robots.

This paper focuses on the task of grape bunch segmentation, which is a critical prerequisite for autonomous plant phenotyping and yield forecast in vineyards  \cite{cecotti_grape_2020,aguiar_grape_2021,santos2020grape}. Our research investigates whether surgical fine-tuning can support grape bunch segmentation under visual domain shifts. To address this research question, we extend the study of surgical fine-tuning from image classification models to instance segmentation architectures in the specific case of viticulture. The work in \cite{cecotti_grape_2020} is most closely related to this study, because it evaluates the utility of linear probing for grape segmentation. However, the experiments in \cite{cecotti_grape_2020} did not examine the option of fine-tuning layers other than the classification head.   

To facilitate the analysis of different types of visual domain shifts that characterise vineyards, we introduce the VINEyard Piacenza Image Collections (VINEPICs) \cite{vinepics}, a comprehensive and novel grape image archive. In \cite{santos2020grape}, Santos et al. presented the Embrapa Wine Grape Instance Segmentation Dataset (WGISD), which is a large-scale collection of vineyard images displaying high-resolution instances of grape bunches across five different grapevine varieties. Our dataset was gathered in a distinct geographic area and it encompasses different grapevine varieties from those in the WGISD dataset, including wine and table grapes. Crucially, the proposed VINEPICs dataset contains additional variations in terms of camera viewpoint, scene occlusion, and time of data collection. Moreover, we captured images using a consumer-grade camera mounted on a mobile robot, which presents additional challenges due to possible motion blur from the robot's movement. As such, the contributed dataset more closely resembles realistic setups in autonomous vineyard phenotyping compared to the WGISD benchmark.

Our results from applying the widely-adopted Mask R-CNN model \cite{he2017mask,he2019rethinking,ghiasi2021simple} to challenging robot-collected images indicate that adopting a surgical fine-tuning strategy can significantly outperform both linear probing and full parameter tuning when novel samples that introduce distribution shifts are considered. The paper is structured as follows. In Section \ref{sec:methods}, we present the reference datasets, ablation study, technical implementation, and evaluation metrics used in our experiments. We then discuss the experimental results in Section \ref{sec:results}. Concluding remarks and future extensions of this work are left to Section \ref{sec:end}.

\section{Materials and methods}\label{sec:methods}

To test the performance of applying surgical fine-tuning to instance segmentation models, we ran a set of layered experiments. Consistently with \cite{lee2023surgical}, we set up the training in two stages. First, we pre-trained on the largest available set of examples for the grape segmentation task: namely WGISD in this case \cite{wgisd2019data}. Then, we considered different target sets that introduce a distribution shift from the source set. The goal was evaluating the extent to which transfer learning can be achieved from source to target, with minimal adjustments, thanks to surgical fine-tuning. 
Differently from \cite{lee2023surgical}, where the evaluation set was held out from the same data used for fine-tuning, we ran inferences on a different dataset, collected one year after the fine-tuning set. This setup resembles the real-world challenges of viticulture applications. Indeed, grape images can be collected only at specific times of the year and adapting learning models from past years to newly-collected data becomes essential.

\subsection{Datasets}
\textbf{Embrapa WGISD.} The Embrapa Wine Grape Instance Segmentation Dataset (WGISD) \cite{wgisd2019data} comprises 300 high-resolution images depicting 2,020 grape bunches from five \textit{Vitis vinifera} L. grapevine varieties: Chardonnay, Cabernet Franc, Cabernet Sauvignon, Sauvignon Blanc, and Syrah. The images were captured at the Guaspari Winery (Espírito Santo do Pinhal, São Paulo, Brazil) in April 2018, with the exception of images of the Syrah dataset that was collected in April 2017. Grape bunches were photographed while keeping the camera principal axis approximately perpendicular to the vineyard row, using both a Canon EOS REBEL T3i DSLR camera and a Motorola Z2 Play smartphone and were resized and stored at a resolution of 2048x1365. At the time of data collection, no defoliation treatments were applied except for the routine canopy management for wine production adopted in the region. In the original data split used in \cite{santos2020grape}, 110 images (accounting for 1612 grape instances) were jointly devoted to training and validation, whereas 27 images (i.e., 408 grape instances) were held out for testing. However, the actual split between training and validation was not provided. Therefore, we decided to use a 20\% validation split stratified across grape varieties from the original training subset.

\begin{table}[t]
\caption{Domain shifts from source to target data.}
\centering
\begin{tabular}{llll}
\toprule
\textbf{Dataset} & \textbf{Changes introduced} & \textbf{Shift types\cite{lee2023surgical}} & \textbf{Instances} \\
\midrule
\textbf{Source:} WGISD & -  & - & 2,020\\ 
 & & & \\
\hline
& geographic area, & natural,\\
\textbf{Fine-tuning set:} & vineyard, & feature-level & \\
VINEPICs21 & Red Globe & input-level & 668 \\
 & camera setup & & \\
\hline
 &  & & \\
\textbf{Target sets:} & & & \\
VINEPICs22R & temporal: & input-level & 100\\
 & different years & &\\
& & & \\
\cline{2-4}
 &  & & \\
VINEPICs22RV & temporal, & input-level & 112 \\
& camera viewpoint & & \\
& & & \\
\cline{2-4}
 &  & \\
VINEPICs22RF & temporal, & input-level & 105 \\
& foliage occlusion & & \\
& & & \\
\cline{2-4}
 &  & & \\
VINEPICs22C & temporal, & input-level & 138 \\
& grape variety  & feature-level & \\
& (Cabernet S.: red) & & \\
& & & \\
\cline{2-4}
 &  & & \\
VINEPICs22O & temporal, & input-level & 135 \\
& grape variety  & feature-level & \\
& (Ortrugo: white) & & \\
&   & & \\
\bottomrule
\end{tabular}
\label{tab:domain}
\end{table}

\textbf{VINEPICs.} The VINEyard Piacenza Image Collections (VINEPICs) dataset consists of grape images collected at the vineyard facility of Università Cattolica del Sacro Cuore in Piacenza, Italy. The VINEPICs dataset is publicly available under CC BY 4.0 (Attribution 4.0 International) license and accessible at this link \url{https://doi.org/10.5281/zenodo.7866442}. The acronym \textbf{VINEPICs21} refers to the first collection of images gathered in the summer of 2021 on Red Globe vines (\textit{Vitis vinifera }L.) grafted on  Selection Oppenheim 4 (SO4), i.e., the vine rootstock, growing outdoors in 25 L pots. This set includes 73 RGB images captured on three different dates: 26 images of resolution 480x848 were collected at beginning of grape ripening on July 27th, 23 images of resolution 720x1280 on August 23rd when berries were fully coloured, and 24 images of resolution 720x1080 at harvest on September 9th. An Intel D435i RGB-D camera was used to capture the data, which was mounted on a SCOUT 2.0 AgileX robotic platform, a four-wheeled differential steering mobile robot\footnote{The analyses presented in this paper only concern RGB images, but we also collected depth data to support a wider range of applications, such as, e.g., estimating the volume of grape bunches.}. The plants were arranged along two, vertically shoot-positioned, North-South oriented rows and hedgerow-trained for a canopy wall extending about 1.3 m above the main wire. Each vine had a $\sim$1 m cane bearing 10-11 nodes that was raised 80 cm from the ground. Between fruit-set (BBCH 71) and berry touch (BBCH 79) \cite{lorenz1995growth}, the leaves around bunches were gradually removed for a resulting fully defoliated fruit zone with reduced incidence of berry sunburns \cite{gatti2015interactions}. Before veraison, eight vines were subjected to crop thinning to control for fruit occlusions caused by excessive fruit density. Accordingly, a basal bunch was kept every second shoot for about six retained bunches/vine; the remaining unthinned vines were clustered into two groups with about 10 and 4 bunches/vine. During data collection, the camera principal axis was rotated to form an angle of approximately \SI{45}{\degree} with the scanned plant row. The grape bunch regions were annotated using polygonal masks through the Computer Vision Annotation Tool (CVAT)\footnote{\url{https://github.com/opencv/cvat}}, and the annotations followed the COCO annotation format\footnote{\url{https://cocodataset.org/}}. \\
A second and more extensive dataset, named \textbf{VINEPICs22}, was collected at the same vineyard facility of Università Cattolica del Sacro Cuore in Piacenza, Italy, on two separate dates in August and September 2022, approximately one year after the previous set. This dataset comprises 165 annotated images, representative of different types of domain shifts, including 1464 grape bunch instances. From this dataset, we extracted subsets of data to control for the incremental changes we expect from the fine-tuning domain (VINEPICs21) to the target domain, as detailed in Table \ref{tab:domain}. Specifically, the VINEPICs22R set includes new images collected from the same grape variety (Red Globe), by maintaining the same camera viewpoint, and level of defoliation as VINEPICs21. VINEPICs22RV introduces a change in the camera viewpoint (i.e., the camera principal axis is perpendicular to the plant rows), while set VINEPICs22RF was captured first on non-defoliated canopies. Furthermore, sets VINEPICs22C and VINEPICs22O maintain the same camera viewpoint and defoliation level as VINEPICs21 but represent different grape varieties, namely Cabernet Sauvignon (red grape) and Ortrugo (white grape), growing in a experimental vineyard. Table \ref{tab:domain} maps the changes introduced for each fine-tuning and target set to the taxonomy of shift types adopted in \cite{lee2023surgical}. The selected target sets cover three shift types: i) \textit{input-level shifts}, which occur due to variations in the visual appearance of the same environment (e.g., observing the same vineyard on different days introduces lighting variations); ii) \textit{feature-level shifts}, where the source-target shift is caused by different populations of the same class, in our case, different grape varieties; and iii) \textit{natural shifts}, which are due to collecting the source and target data in different environments, in our case, different growing conditions (potted vines vs. experimental vineyard). \textit{Output-level shifts} do not concern our use-case, since the target class (grape bunches) remains unchanged throughout the experiments detailed in this paper.

\begin{figure*}[t!]
  \includegraphics[width=0.9\textwidth,trim=0 0 0 0, clip]{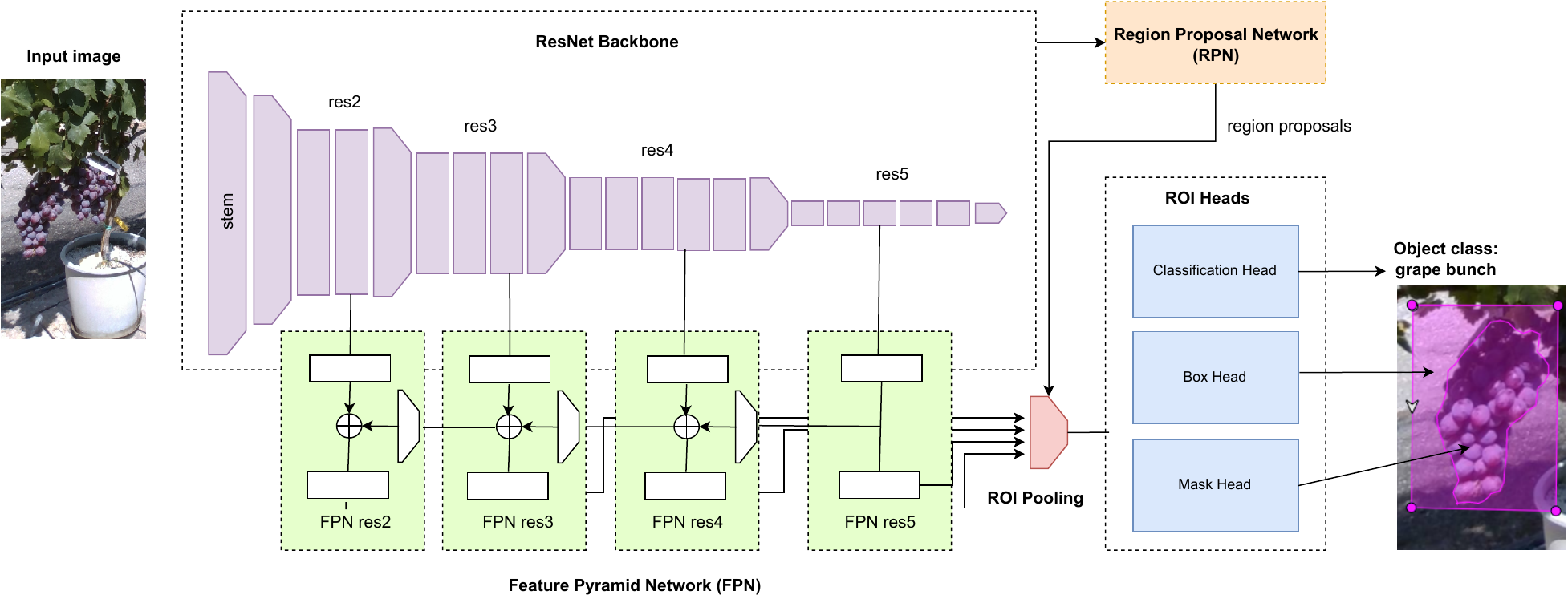}
\caption{Overview of the Mask R-CNN architecture. The backbone of the architecture is based on ResNet50, and features from blocks 2 to 5 are extracted and passed through a Feature Pyramid Network (FPN). The Region Proposal Network (RPN) generates region proposals, which are then combined with the upsampled features and input to three model heads, which predict object class, bounding box, and polygonal mask in parallel.}
  \label{fig:arch} 
\end{figure*}

\subsection{Surgical fine-tuning for instance segmentation}\label{sec:method}
Given the focus on image classification tasks, the experiments described in \cite{lee2023surgical} consider ResNet architectures \cite{he2016ResNet} as a reference and utilize surgical fine-tuning to manipulate the different residual blocks. However, in the context of instance segmentation tasks, supplementary modules are introduced for detecting and segmenting object regions. Region-based segmentation architectures such as the widely utilized Mask R-CNN model \cite{he2017mask} merge CNN feature extraction layers with a Region Proposal Network (RPN) that extracts Regions of Interest (ROI) from input images. Predicted object regions are then fed to three network heads that operate in parallel, generating predictions for the object class, bounding box, and polygonal mask (Figure \ref{fig:arch}). A popular implementation of this generalized architecture uses a combination of ResNets and Feature Pyramid Networks (FPN) as a backbone for the feature extraction step \cite{he2019rethinking,ghiasi2021simple}.

To assess the efficacy of surgical fine-tuning in the context of region-based segmentation models, we also ought to examine the impact of selectively fine-tuning the FPN and RPN components, along with the residual blocks and classification heads. Hence, we conduct experiments that compare the following model ablations:

\begin{itemize}
\item{\textbf{Tune All:}} This configuration fine-tunes all model parameters.
\item{\textbf{Linear Probing:}} In this classic configuration, only parameters in the three ROI heads are updated, while earlier layer parameters remain fixed at values learned during pre-training.
\item{\textbf{Res n:}} This setup involves fine-tuning only the ResNet layers, specifically the residual block identified by the number $n$. We use the keyword "stem" to refer to the first residual block, and the notation "res n" for blocks numbered 2 and higher. This setup follows the rationale applied in \cite{lee2023surgical}.
\item{\textbf{Joint SFT: Res Block n + FPN at n:}} This configuration is a variation of the previous setup, where the selected residual blocks are fine-tuned simultaneously with the related Feature Pyramid Network (FPN) operations.
\item{\textbf{RPN:}} In this setup, we only apply surgical fine-tuning to the Region Proposal Network (RPN) in the Mask R-CNN model.
\end{itemize}

To the best of our knowledge, this is the first study on the application of surgical fine-tuning to instance segmentation tasks.  

\begin{table*}[t]
\caption{Inference results from pre-training baseline instance segmentation models on the WGISD dataset.}
\centering
\begin{tabular}{l|c|c|c|c}
\toprule
\textbf{Baseline} & \textbf{AP}$_{0.3-0.9}$ & \textbf{P}$_{0.3-0.9}$ & \textbf{R}$_{0.3-0.9}$ & \textbf{F1}$_{0.3-0.9}$ \\
\midrule
Mask R-CNN ResNet101 (results from \cite{santos2020grape}) & 0.540 & 0.683 & 0.649 & 0.665 \\
Mask R-CNN ResNet101 \cite{ghiasi2021simple} & 0.550 &  0.789 & 0.588 & 0.674 \\
Mask R-CNN ResNet50 \cite{ghiasi2021simple} & 0.571 & \textbf{0.806} & 0.607 & 0.693\\
Mask R-CNN ResNet50 \cite{he2019rethinking} & \textbf{0.623}& 0.796 & \textbf{0.663}  & \textbf{0.724} \\
\bottomrule
\end{tabular}
\label{tab:source}
\end{table*}

\begin{table}[htbp]
\caption{Number of fine-tuned parameters in the evaluated ablations.}
\centering
\begin{tabular}{l|l}
\toprule
\textbf{Ablation} & \textbf{Parameters } \\
\midrule
tune all & $\thicksim$ 45.3M\\
 linear probing & $\thicksim$ 17.8M\\
 stem & $\thicksim$ 9.5K \\
 res2 & $\thicksim$ 215K\\
 res2 + FPN & $\thicksim$ 872K \\
 res3 & $\thicksim$ 1.22M\\
 res3 + FPN & $\thicksim$ 1.94M \\
 res4 & $\thicksim$ 7.1M \\
 res4 + FPN & $\thicksim$ 7.95M \\
 res5 & $\thicksim$ 14.9M \\
 res5 + FPN & $\thicksim$ 16.1M\\
 RPN & $\thicksim$ 594K\\
\bottomrule
\end{tabular}
\label{tab:params}
\end{table}

\subsection{Implementation details}
To apply surgical fine-tuning as described in the previous section, we customised the Detectron2\footnote{\url{https://github.com/facebookresearch/detectron2}} implementation of the Mask R-CNN architecture. The code for reproducing these trials is available at \url{https://github.com/AIRLab-POLIMI/SFT_grape_segmentation}. 

We augmented our training examples by applying various transformations such as Gaussian blur, additive Gaussian noise, random brightness, contrast, and saturation, pixel dropout, and random flipping transformations. During pre-training on the source domain, we utilized ResNet50 and ResNet101 backbones employing Group Normalization (GN). We experimented with different weight initializations following the Detectron2 Mask R-CNN baselines for the COCO instance segmentation task. In the first configuration, we used the weights obtained from the method introduced in \cite{ghiasi2021simple}, where the model was trained from scratch on COCO with an extended training schedule and an augmented jittering scale. In the second configuration, we initialized the model with the weights from the method presented in \cite{he2019rethinking}, where Mask R-CNN was trained on COCO instances from scratch, i.e., with random weight initialization, rather than reusing initialization values derived from ImageNet. All models were trained with a batch size of 2 images, and we used an early stopping criterion if the validation loss did not improve for 30 consecutive evaluation checks, with one evaluation check every 220 minibatch iterations. We optimized model parameters using stochastic gradient descent, with a constant learning rate set to 0.01.

\subsection{Evaluation metrics}
We evaluate the instance segmentation performance by measuring the Average Precision (AP) of predicted object regions, as well as the standard Precision (P), Recall (R), and F1 score of predicted object instances. The metrics were averaged over Intersection over Union (IoU) values ranging from 0.3 to 0.9, to allow for comparison with the results presented in \cite{santos2020grape}. Consistently with \cite{santos2020grape}, only predictions with confidence greater than 0.9 for the grape class are considered in the evaluation. 
We prioritize improvements in terms of F1 over individual P and R scores, as detecting all true positives is as important as minimizing the false positives in the target use-case.

\begin{figure*}
\captionsetup[subfigure]{justification=centering}
     \begin{subfigure}[b]{0.1\textwidth}
        \centering
         \includegraphics[scale=0.18, trim=380 0 0 0 0, clip]{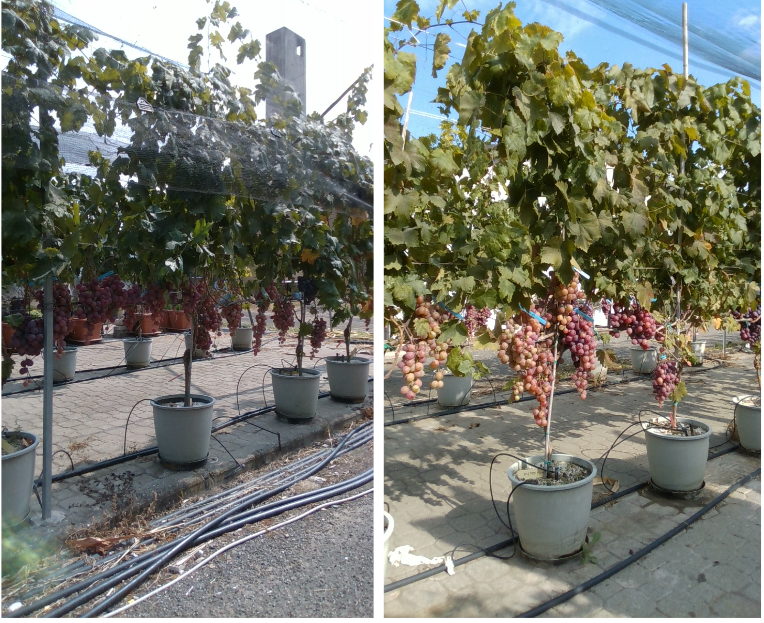}
         \caption{VINEPICs21}
         \label{fig:cat21}
     \end{subfigure}
     \hfill
     \begin{subfigure}[b]{0.1\textwidth}
         \centering
         \includegraphics[scale=0.2, trim=325 0 0 0 0, clip]{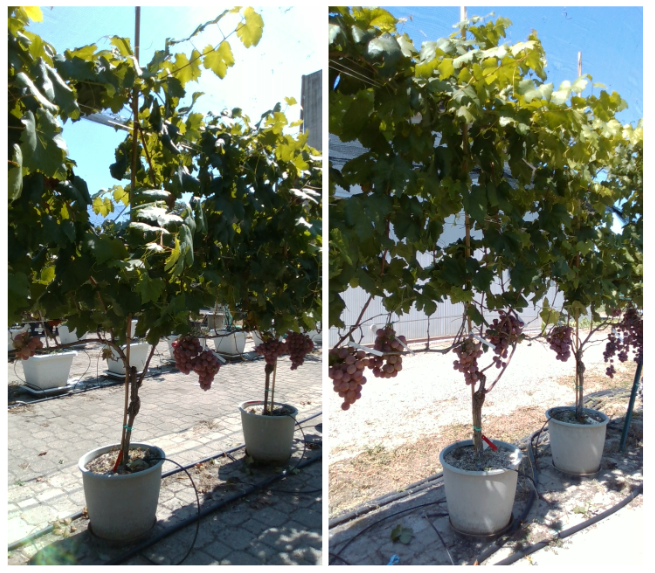}
         \caption{VINEPICs22R}
         \label{fig:cat22A}
     \end{subfigure}
     \hfill
     \begin{subfigure}[b]{0.1\textwidth}
         \centering
         \includegraphics[scale=0.2, trim=320 0 0 0 0, clip]{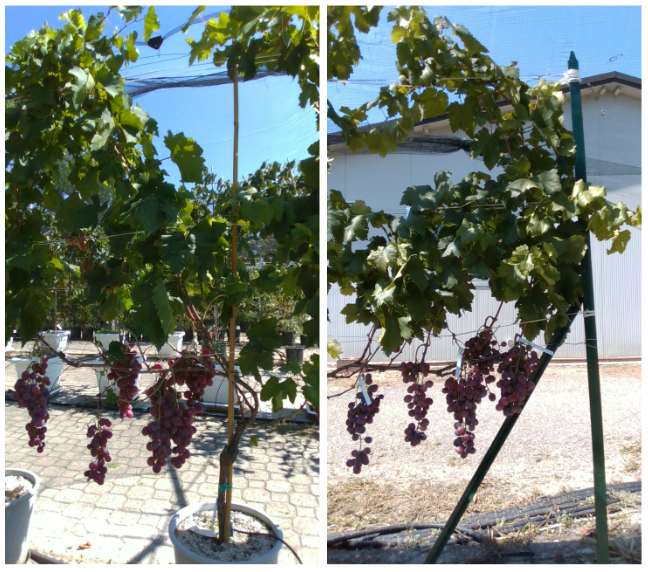}
         \caption{VINEPICs22RV}
         \label{fig:cat22B}
     \end{subfigure}
     \hfill
     \begin{subfigure}[b]{0.1\textwidth}
         \centering
         \includegraphics[scale=0.2, trim=325 0 0 0 0, clip]{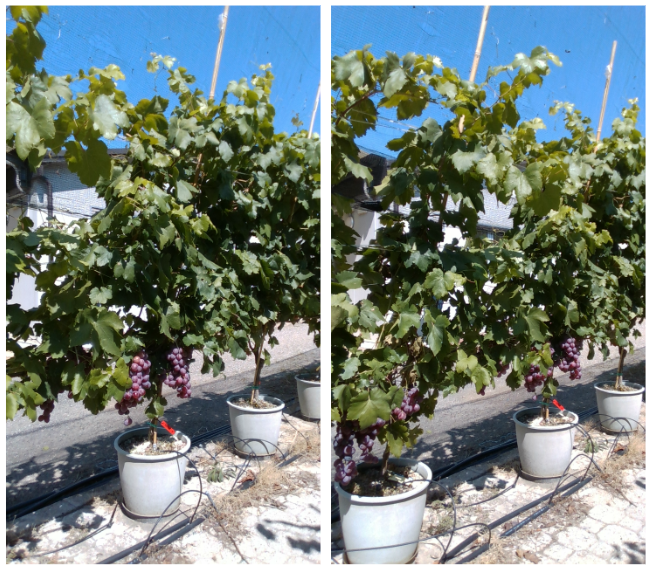}
         \caption{VINEPICs22RF}
         \label{fig:cat22C}
     \end{subfigure}
    \hfill
     \begin{subfigure}[b]{0.1\textwidth}
         \centering
         \includegraphics[scale=0.2, trim=325 0 0 0 0, clip]{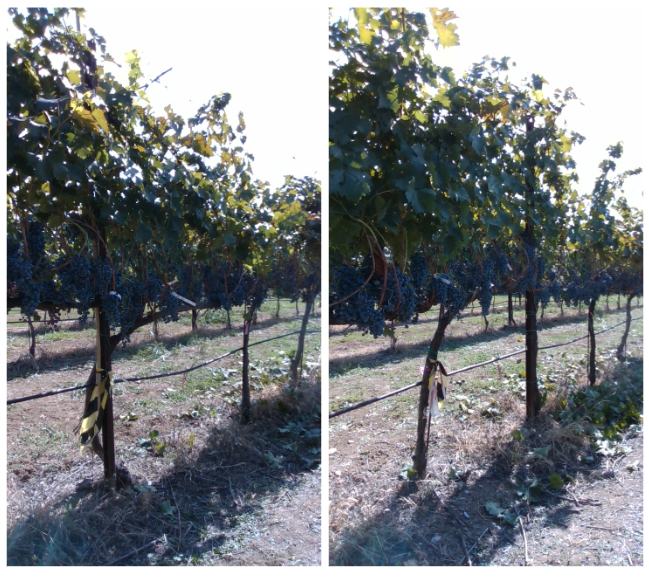}
         \caption{VINEPICs22C}
         \label{fig:cat22D}
     \end{subfigure}
    \hfill
     \begin{subfigure}[b]{0.1\textwidth}
         \centering
         \includegraphics[scale=0.2, trim=325 0 0 0 0, clip]{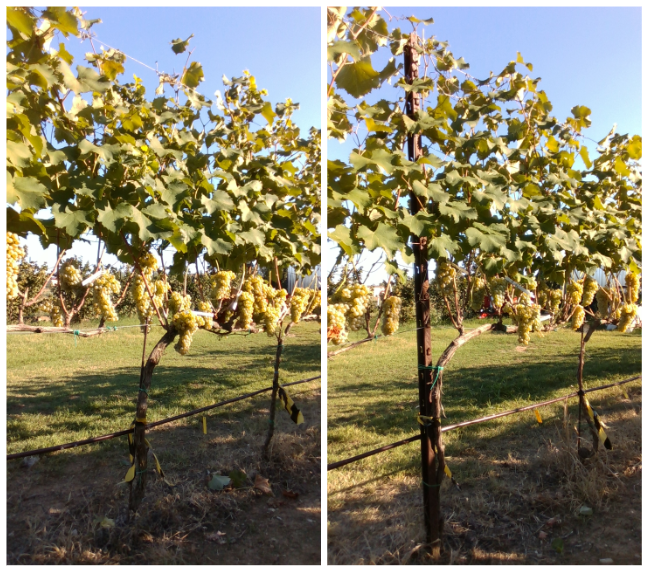}
         \caption{VINEPICs22O}
         \label{fig:cat22E}
     \end{subfigure}
\caption{Image examples from the VINEPICs sets. Examples from the WGIS set are available in \cite{wgisd2019data}.}
        \label{fig:examples}
\end{figure*}

\begin{table*}[t!]
\centering
\caption{Inference results on test sets, after applying surgical fine-tuning on VINEPICs21. }
\begin{tabular}{l|l|ccccc}
\toprule
\textbf{Test set}& \textbf{Ablations} & \textbf{AP}$_{0.3-0.9}$ & \textbf{P}$_{0.3-0.9}$ & \textbf{R}$_{0.3-0.9}$ & \textbf{F1}$_{0.3-0.9}$ \\
\midrule
VINEPICs21 test & tune all & 0.374 & 0.767 & 0.404 & 0.529\\
\hline 
 \multirow{3}{*}{VINEPICs22R} & tune all & 0.254  & 0.682 & 0.273 & 0.390 \\  
 & linear probing & 0.226 & \textbf{0.689} & 0.234 & 0.350 \\
& \textbf{res3} & \textbf{0.395} & 0.602 & \textbf{0.421} & \textbf{0.496} \\ 
\hline
\multirow{3}{*}{VINEPICs22RV} & tune all & 0.387  & 0.634 & 0.436 & 0.517\\ 
 & linear probing & 0.409 &  0.660  & 0.454 & 0.538 \\ 
& \textbf{res4 + FPN} & \textbf{0.463}  & 0.595 & \textbf{0.515} & \textbf{0.552}\\
& RPN  & 0.305 & \textbf{0.687} &0.325  & 0.442\\
\hline
\multirow{3}{*}{VINEPICs22RF} & tune all & 0.342 & 0.696 & 0.371 & 0.484\\ 
 & linear probing & 0.290 & \textbf{0.711} & 0.305 & 0.426\\
& \textbf{res3}  & \textbf{0.469} &  0.577 & \textbf{0.512}  & 0.542 \\
& \textbf{res3 + FPN} & 0.461  & 0.607 & 0.503 & \textbf{0.550}\\
\hline
\multirow{3}{*}{VINEPICs22C} & tune all & 0.007 & 0.571   & 0.004 & 0.008 \\  
 & linear probing & 0.003 & 0.286  & 0.002 & 0.004\\ 
& res2 &  0.013 & \textbf{0.643} & 0.009 & 0.018\\
& \textbf{res4}  & \textbf{0.068} & 0.534 & \textbf{0.073} & \textbf{0.129}\\
& res4 + FPN & \textbf{0.068} & 0.548 & 0.071 & 0.126\\
\hline
\multirow{3}{*}{VINEPICs22O}& tune all &  0.022 & \textbf{0.762} & 0.017 & 0.033\\ 
 & linear probing & 0.021  & 0.449 & 0.023 & 0.044\\ 
& \textbf{res4 + FPN} & \textbf{0.102} & 0.625 & \textbf{0.111} & \textbf{0.189}\\
\bottomrule
\end{tabular}
\label{tab:target_small}
\end{table*}

\section{Results and Discussion}\label{sec:results}
Before conducting the ablation study, we pre-trained three Mask R-CNN models on the WGISD dataset. Table \ref{tab:source} demonstrates that on our task, ResNet50 backbones generally delivered better results than ResNet101 backbones. Furthermore, initializing the model with weights obtained after training from scratch on the COCO dataset \cite{he2019rethinking} yielded the best combination of segmented object region quality (in terms of AP) and grape class prediction quality (in terms of F1), compared to using weights from longer training schedules and large-scale jittering \cite{ghiasi2021simple}. Therefore, we have chosen the "Mask R-CNN ResNet50 \cite{he2019rethinking}" model as the baseline for fine-tuning on VINEPICs21.

During the fine-tuning stage, we applied the different ablations presented in Section \ref{sec:method} and evaluated the results on the five target sets selected from VINEPICs22. The top-performing methods in each set of trials, together with the "linear probing" and "tune all" alternatives, are summarised in Table \ref{tab:target_small}. The complete evaluation results can be found in the appendix of this paper (Table \ref{table:target}). We also report the number of parameters tuned in each configuration in Table \ref{tab:params}. 

Results on the \textbf{VINEPICs22R} sets approximate scenarios where the only change introduced is the date and time of data collection, while considering the same grape variety (Red Globe), camera viewpoint, and defoliation level as the fine-tuning set. In this case, fine-tuning the first four CNN layers individually, excluding the stem, ensured a higher AP than the scenario when all model parameters are tuned. In particular, tuning the third ResNet block led to the highest AP and F1 scores, outperforming linear probing. 

Changing camera viewpoint, in \textbf{VINEPICs22RV}, led to generally higher scores than the previous set of trials. Notably, the AP scores are even higher than the AP achieved on the VINEPICs21 test set, for the majority of tested ablations. This result may be due to the fact that a perpendicular camera viewpoint is more similar to the setup adopted in the WGISD set, i.e., the source set. Moreover, it is worth noting that the VINEPICs21 test split comprises nearly twice as many grape instances as the VINEPICs22RV set. As a result, the average scores in the VINEPICs21 case provide more conservative performance figures than VINEPICs22, which accounts for approximately 100 instances for each subset (Table \ref{tab:domain}). In this case, tuning the third and fourth ResNet blocks led to the most marked improvement over the the "tune all" and "linear probing" performance. In particular, tuning the fourth ResNet block in combination with its FPN layers led to the highest results with respect to the AP of region predictions, Recall and F1 of instance predictions. Interestingly, the top precision was achieved when tuning the Region Proposal Network in isolation, albeit generating a higher number of false positives, as indicated by the lower recall scores.  

We then considered grape images captured in the presence of occluding foliage (\textbf{VINEPICs22RF}), under temporal and viewpoint conditions that are comparable to the tuning set. Similarly to the case of the temporal shifts introduced in VINEPICs22R, the top performance was achieved by tuning the third ResNet block. However, in this case, while the highest AP score was achieved in the "res3" configuration, the highest F1 was reached by jointly tuning res3 with FPN.

When we shift the target domain towards different grape varieties, the drop in performance from the fine-tuning set to the target sets is significant. Indeed, although the source set (WGISD) already included examples of both red and white grape bunches, the VINEPICs22C and VINEPICs22O sets are drastically more challenging than previously examined sets. First, the number of instances to be detected in each frame is significantly higher in this case, as exemplified in Figure \ref{fig:examples}. Moreover, images in these sets were captured at a lower resolution than WGISD and in lower lighting conditions than both the WGISD and the VINEPICs21 sets. Thus, this setup complicates not only the learning but also the manual annotation of grape instances. Under these challenging conditions, selectively tuning the stem and RPN was ineffective and prevented the model from providing any grape predictions (Table \ref{table:target}). 
Conversely, applying surgical fine-tuning to intermediate layers resulted in a significant improvement over the near-zero baseline performance. In the case of the Cabernet Sauvignon variety (\textbf{VINEPICs22C}) tuning only the parameters in the fourth ResNet block improved the AP by 10\% and the F1 by 12\%, compared to "linear probing". In the case of the Ortrugo variety (\textbf{VINEPICs22O}), jointly tuning res4 with FPN outperformed "linear probing" by 8\%, in terms of AP, and by 14\%, in terms of F1.  

Overall, results from these experiments support the view that selecting intermediate network layers can outperform the common practice of only re-training the classification head of the model, when visual domain shifts are introduced. In particular, we found that selecting the third block for fine-tuning best supported temporal changes, as well as changes in the level of plant defoliation. Selecting the fourth ResNet block, instead, contributed to mitigating the impact of viewpoint and grape variety shifts. 
Importantly, adopting a surgical fine-tuning approach allowed us to substantially reduce the number of parameter updates, compared to the costly alternative of re-training the complete model from scratch: from over 45M total parameters to nearly 1M and 7M in the res3 and res4 cases (Table \ref{tab:params}).

\section{CONCLUSIONS}\label{sec:end}
To effectively deploy mobile robots for agricultural applications, improving the adaptability of visual perception methods based on Deep Learning to rapidly-changing environments is essential. In particular, we have considered the task of autonomously segmenting grape instances from images collected in real vineyards. In this context, we showed that pre-training on large-scale, high-resolution training examples and fine-tuning only selected layers on more challenging robot-collected data can support knowledge transfer to newly-collected grape images that introduce changes in the camera viewpoint, foliage occlusion level, and grape variety. 

Notably, tuning intermediate network layers improves the robustness of the model to input-level and feature-level shifts. These findings complement the evidence gathered in \cite{lee2023surgical} on image classification benchmarks, where input-level shifts were best supported by tuning the initial network layers. These results also withstand the popular practice of only tuning the last layers on a new target domain. Even in challenging scenarios where images of novel grape varieties are introduced at test time, surgical fine-tuning on intermediate network blocks allowed us to bootstrap the grape segmentation performance, while drastically reducing the number of parameters required for fine-tuning.  

Our evaluation of the utility of surgical fine-tuning to support grape segmentation has been limited to methods derived from the widely-applied Mask R-CNN architecture. Thus, future research directions include the study of instance segmentation models that are based on Transformers, such as \cite{fang2022eva}, for instance. Another transfer learning approach that we have not yet explored concerns the combination of linear probing with the selection of useful features from different layers, as proposed in \cite{evci2022head2toe}.      
The availability of the VINEPICs resource can facilitate the progress in tackling these unexplored research directions. 


\section*{APPENDIX}
Table \ref{table:target} reports the complete evaluation results for the VINEPICs22 target sets.

\section*{ACKNOWLEDGMENTS}
This paper is supported by the Italian L’Oreal-UNESCO program “For Women in Science”, the European Union's
Digital Europe Programme under grant agreement Nº 101100622 (AgrifoodTEF). The study was conducted within the Agritech National Research Center and received funding from the European Union Next-GenerationEU (PIANO NAZIONALE DI RIPRESA E RESILIENZA (PNRR) – MISSIONE 4 COMPONENTE 2, INVESTIMENTO 1.4 – D.D. 1032 17/06/2022, CN00000022).



\bibliographystyle{IEEEtran}
\bibliography{agri}

\begin{table*}[htbp]
\caption{Inference results on test sets, after applying surgical fine-tuning on VINEPICs21.}
\centering
\begin{tabular}{l|l|ccccc}
\toprule
\textbf{Test set}& \textbf{Ablations} & \textbf{AP}$_{0.3-0.9}$ & \textbf{P}$_{0.3-0.9}$ & \textbf{R}$_{0.3-0.9}$ & \textbf{F1}$_{0.3-0.9}$ \\
\midrule
VINEPICs21 test & tune all & 0.374 & 0.767 & 0.404 & 0.529\\
\hline 
 \multirow{12}{*}{VINEPICs22R} & tune all & 0.254  & 0.682 & 0.273 & 0.390 \\  
 & linear probing & 0.226 & \textbf{0.689} & 0.234 & 0.350 \\
 & stem & 0.101 &  0.579  & 0.104 & 0.177\\
& res2 & 0.296 & 0.626 & 0.313  & 0.417\\
& res2 + FPN & 0.347 & 0.638 & 0.37 & 0.468 \\ 
& \textbf{res3} & \textbf{0.395} & 0.602 & \textbf{0.421} & \textbf{0.496} \\ 
& res3 + FPN & 0.389 & 0.607 & 0.407 & 0.488 \\
& res4 & 0.277 & 0.613 & 0.294 & 0.398 \\ 
& res4 + FPN & 0.338 & 0.571 &  0.36 & 0.442\\ 
& res5  & 0.226 & 0.619  & 0.241 & 0.347 \\ 
& res5 + FPN & 0.222 & 0.624 & 0.237  & 0.343 \\
& RPN  &  0.009 & 0.185 & 0.031 & 0.053 \\
\hline
\multirow{12}{*}{VINEPICs22RV} & tune all & 0.387  & 0.634 & 0.436 & 0.517\\ 
 & linear probing & 0.409 &  0.660  & 0.454 & 0.538 \\ 
& stem & 0.391 & 0.681 &  0.420 & 0.519 \\
& res2 & 0.392 & 0.615 &  0.450  & 0.520\\
& res2 + FPN &  0.447 & 0.621 & 0.493 & 0.550\\
& res3  &0.440 & 0.555 & 0.510  & 0.531\\
& res3 + FPN & 0.437 & 0.567 & 0.511 & 0.538\\
& res4  & 0.430 & 0.584 &  0.475  & 0.524\\
& res4 + FPN & \textbf{0.463}  & 0.595 & \textbf{0.515} & \textbf{0.552}\\
& res5  & 0.412 & 0.604 &  0.448  & 0.514\\
& res5 + FPN & 0.391  & 0.625 & 0.430 & 0.510\\
& RPN  & 0.305 & \textbf{0.687} &0.325  & 0.442\\
\hline
\multirow{12}{*}{VINEPICs22RF} & tune all & 0.342 & 0.696 & 0.371 & 0.484\\ 
 & linear probing & 0.290 & \textbf{0.711} & 0.305 & 0.426\\
& stem & 0.220 & 0.694 & 0.231  & 0.347 \\
& res2 & 0.405 & 0.688 &0.432  & 0.531 \\
& res2 + FPN & 0.415 & 0.626 & 0.453  & 0.525\\
& res3  & \textbf{0.469} &  0.577 & \textbf{0.512}  & 0.542 \\
& res3 + FPN & 0.461  & 0.607 & 0.503 & \textbf{0.550}\\
& res4  & 0.391 & 0.556 & 0.434 & 0.487 \\
& res4 + FPN &  0.423  & 0.566 & 0.475 & 0.516 \\
& res5  & 0.332  & 0.563 & 0.375   & 0.450\\
& res5 + FPN & 0.299 & 0.626   & 0.328 & 0.430 \\
& RPN  & 0.041 & 0.295 & 0.045  & 0.078 \\
\hline
\multirow{12}{*}{VINEPICs22C} & tune all & 0.007 & 0.571   & 0.004 & 0.008 \\  
 & linear probing & 0.003 & 0.286  & 0.002 & 0.004\\ 
& stem &  no & predictions  & from   & model\\
& res2 &  0.013 & \textbf{0.643} & 0.009 & 0.018\\
& res2 + FPN &  0.016 & 0.500 & 0.014 & 0.028 \\
& res3  & 0.040 & 0.543 & 0.039 & 0.073 \\
& res3 + FPN & 0.066 & 0.442 & 0.070 & 0.121\\
& res4  & \textbf{0.068} & 0.534 & \textbf{0.073} & \textbf{0.129}\\
& res4 + FPN & \textbf{0.068} & 0.548 & 0.071 & 0.126\\
& res5  & 0.039 & 0.543 &  0.039  & 0.073 \\
& res5 + FPN &  0.029  & 0.571 & 0.029 & 0.055\\
& RPN  & no & predictions  & from   & model\\
\hline
\multirow{12}{*}{VINEPICs22O}& tune all &  0.022 & \textbf{0.762} & 0.017 & 0.033\\ 
 & linear probing & 0.021  & 0.449 & 0.023 & 0.044\\ 
& stem &  no & predictions  & from   & model\\
& res2 & 0.017 & 0.428  & 0.013 & 0.025\\
& res2 + FPN &  0.045 & 0.614 & 0.045 & 0.085 \\
& res3  & 0.047 & 0.527 & 0.051 & 0.093\\
& res3 + FPN & 0.071  & 0.571 & 0.076 & 0.134\\
& res4  & 0.070 & 0.676 &  0.075  & 0.135\\
& res4 + FPN & \textbf{0.102} & 0.625 & \textbf{0.111} & \textbf{0.189}\\
& res5  & 0.059 & 0.458 &  0.064  & 0.113\\
& res5 + FPN & 0.077 & 0.661 & 0.078 & 0.140\\
& RPN  &  no & predictions  & from   & model\\
\bottomrule
\end{tabular}
\label{table:target}
\end{table*}

\end{document}